\documentclass{article}
\usepackage{spconf,amsmath,graphicx,booktabs,cite,amsfonts,verbatim,multirow,makecell}

\title{SPEAKER-AWARE SPEECH-TRANSFORMER}
\name{Zhiyun Fan$^{1,2}$, Jie Li$^3$, Shiyu Zhou$^1$, Bo Xu$^1$\thanks{This work is supported by the National Key Research and Development Program of China under No. 2016YFB1001404.}}
\address{
 $^1$Institute of Automation, Chinese Academy of Sciences, China\\
 $^2$University of Chinese Academy of Sciences, China\\
 $^3$Kwai, Beijing, P.R. China\\
\{fanzhiyun2017, zhoushiyu2013, xubo\}@ia.ac.cn, lijie03@kuaishou.com}

\begin{document}
%
\maketitle
\begin{abstract}
Recently, end-to-end (E2E) models become a competitive alternative to the conventional hybrid automatic speech recognition (ASR) systems. However, they still suffer from speaker mismatch in training and testing condition. In this paper, we use Speech-Transformer (ST) as the study platform to investigate speaker aware training of E2E models. We propose a model called Speaker-Aware Speech-Transformer (SAST), which is a standard ST equipped with a speaker attention module (SAM). The SAM has a static speaker knowledge block (SKB) that is made of i-vectors. At each time step, the encoder output attends to the i-vectors in the block, and generates a weighted combined speaker embedding vector, which helps the model to normalize the speaker variations. The SAST model trained in this way becomes independent of specific training speakers and thus generalizes better to unseen testing speakers. We investigate different factors of SAM. Experimental results on the AISHELL-1 task show that SAST achieves a relative $6.5\%$ CER reduction (CERR) over the speaker-independent (SI) baseline. Moreover, we demonstrate that SAST still works quite well even if the i-vectors in SKB all come from a different data source other than the acoustic training set.

\end{abstract}
\section{Introduction}
\label{sec:intro}

There have been growing interests in building an E2E speech recognition system, which directly transduces an input sequence of acoustic features to an output sequence of tokens. Comparing to a conventional hybrid system, such an E2E system typically has several advantages, including a simpler training process, allowing a joint optimization among components, and a compact model size. Prominent E2E approaches include: (a) connectionist temporal classification (CTC) \cite{graves2006connectionist}\cite{graves2014towards}, (b) attention based encoder-decoder networks \cite{cho2014learning, bahdanau2014neural, bahdanau2016end, chorowski2015attention, vaswani2017attention}, and (c) recurrent neural network transducer (RNN-T) \cite{graves2012sequence}. The above approaches have been successfully applied to large-scale ASR tasks \cite{DBLP:journals/corr/abs-1901-01239, dong2018speech, yuanyuan2019speech, zhou2018comparison, Zhou2018}.

Speaker adaptation is an essential component of state-of-the-art hybrid systems, and a variety of adaptation methods have been developed, which can be divided into three categories \cite{DBLP:journals/corr/abs-1901-01239}. The first kind of method, also the most straightforward one, is model re-training. The certain layers or even the whole layers of speaker independent (SI) model are re-updated using the adaptation data of each testing speaker \cite{liao2013speaker, yu2013kl}. To avoid overfitting, L2 regularization \cite{liao2013speaker} and Kullback-Leibler divergence (KLD) regularization \cite{yu2013kl} were applied. The second category is transformation based. Speaker dependent (SD) transformations were used to convert a SI model to a SD model \cite{DBLP:journals/corr/abs-1901-01239}. The third one is speaker-aware training. The acoustic models are trained with speaker auxiliary vectors, such as i-vectors \cite{saon2013speaker, miao2014towards}, speaker code \cite{abdel2013fast}, and speaker embedding \cite{DBLP:journals/corr/abs-1710-06937}, to facilitate the models to normalize the speaker variations.

The aforementioned adaptation methods are mainly investigated in the hybrid systems. Their effectiveness for E2E systems is still not fully studied. In this work, we focus on the speaker-aware training (the third category of method mentioned above) for E2E systems. The study platform that we used is Speech-Transformer (ST), but the methods proposed in this work can be easily generalized to other types of E2E systems, for example, the popular listen, attend and spell (LAS) model \cite{chan2016listen, Changhao2019improving}.

Speech-Transformer, which is based on the Transformer from machine translation task \cite{vaswani2017attention}, was first proposed in \cite{dong2018speech} and then developed in \cite{zhou2018comparison, Zhou2018}. Recently, it was further optimized in \cite{yuanyuan2019speech} and showed a superior performance over a strong hybrid TDNN-LSTM system on a large-scale ASR task. The most straightforward way to do speaker-aware training of ST model is to attach speaker auxiliary vectors (i-vectors in this work) to the input of the model. However, it just brings quite limited performance benefits in our study. It is possibly because the self-attention mechanism makes the encoder more powerful to capture and normalize long-range speaker characteristics.

In this work, we propose a model called Speaker-Aware Speech-Transformer (SAST), which is more effective than simply sending i-vectors to the model directly. SAST contains two parts: the main transformer part and the Speaker Attention Module (SAM). The SAM is trained to generate a soft speaker embedding, which makes the main transformer part become independent of specific training speakers and thus generalize better to unseen testing ones.
We investigate different factors of SAM: the number of i-vectors in SKB, the location of SAM in the model, and the level of soft speaker embedding. Experimental results on the AISHELL-1 task show that SAST achieves a relative $6.5\%$ CER reduction (CERR) over SI baseline. What's more, we show that the speakers in the knowledge block can be different from those used for acoustic model training. This finding further verifies our assumption and provides new clues to the adaptation method of attention-based E2E models.


\section{Related Work}
\label{sec:format}

There have been few efforts on the adaptation of the E2E systems. \cite{ochiai2018speaker} proposed a multi-path adaptation scheme for end-to-end multichannel ASR. In \cite{li2018advancing}, the authors addressed the data sparsity issue by formulating Kullback-Leibler divergence (KLD) regularization and multi-task learning approaches for speaker adaptation of CTC models. The methods investigated in these two works fall in the category of model re-training adaptation, which needs extra training data and parameter storage space for each test speaker. The work \cite{tomashenko2018evaluation} explored different feature-space adaptation approaches for bidirectional long short-term memory (BLSTM)-CTC models. Although some of them are effective, they need an extra second-pass decoding. The authors in \cite{delcroix2018auxiliary} employed a sequence summary network to compute auxiliary features in the input layer of an attention-based E2E model.

The most similar idea to our SAST model is \cite{pan2018online}, where attention mechanism is used to select the most relevant speaker i-vectors to the current speech segment from the memory. The biggest difference lies in the framework: hybrid system in \cite{pan2018online} while E2E in this work. Besides, we investigate different factors of SAM, including the number of i-vectors in SKB, the location of SAM in the encoder, and the level of speaker embedding. We also discuss the difference between the soft and hard speaker embedding.

\section{Background}
\label{sec:bac}

\subsection{Multi-head Attention}
\label{ssec:subhead}
Self-attention, a mechanism that relates different positions of input sequences to compute representations for the inputs. A self-attention layer receives the tuple of query, key and value, and outputs the weighted sum of the value. The weight assigned to each value is computed by a compatibility function of the query with the corresponding key. The outputs of self-attention are computed as:
\begin{equation}
 \setlength{\abovedisplayskip}{4pt}
 \setlength{\belowdisplayskip}{4pt}
  \text{Attention}(Q,K,V) = \text{softmax}(\frac{QK^{T}}{\sqrt{d_k}})V
  \label{eq1}
\end{equation}

A multi-head attention (MHA) layer projects the queries, keys and values to $d_q$, $d_{kv}$ and $d_{kv}$ demensions $h$ times with different trainable projections. And on each version of queries, keys and values, attention function described above is performed in parallel. Then their outputs are concatenated and fed
into another linear projection to obtain the final dimension $d_{model}$ with $W^O$. Formally, the multi-head attention layer is computed as:
\begin{equation}
 \setlength{\abovedisplayskip}{4pt}
 \setlength{\belowdisplayskip}{4pt}
 \text{MHA}(Q,K,V) = \text{Concat}(head_1,..,head_h)W^O
  \label{eq2}
\end{equation}
\begin{equation}
 \setlength{\abovedisplayskip}{4pt}
 \setlength{\belowdisplayskip}{4pt}
   head_i=\text{Attention}(QW_i^Q,KW_i^K,VW_i^V)
  \label{eq3}
\end{equation}
Where the parameter matrices $W_i^Q\in\mathbb{R}^{d_{model} \times  d_q}$, $W_i^K\in\mathbb{R}^{d_{model} \times  d_{kv}}$, $W_i^V\in\mathbb{R}^{d_{model} \times  d_{kv}}$ and $W^O\in\mathbb{R}^{hd_{kv} \times  d_{model}}$, $h$ is the number of heads, and $d_{model}$ is the model dimension.
\subsection{Position-wise Feed-Forward Networks}
\label{ssec:subhead}
A position-wise feed-forward network (FFN) stacks two fullly connected layers with activation function. Different from \cite{vaswani2017attention}, we use Gated Linear Units (GLU) instead of Rectified Linear Unit (ReLU) as activation function. The dimension of input and output is $d_{model}$, and the inner layer has dimension $d_{ff}$.
\begin{equation}
 \setlength{\abovedisplayskip}{4pt}
 \setlength{\belowdisplayskip}{4pt}
  \text{FFN}(x) = GLU(xW_1 + b_1)W_2 + b_2
  \label{eq4}
\end{equation}
Where the weight $W_1 \in \mathbb{R}^{d_{model} \times d_{ff}}$, $W_2 \in \mathbb{R}^{d_{ff} \times d_{model}}$ and the biases $b_1 \in \mathbb{R}^{d_{ff}}$, $b_2 \in \mathbb{R}^{d_{model}}$.

\begin{figure*}[htb]

\begin{minipage}[b]{.48\linewidth}
  \centering
  \centerline{\includegraphics[width=5.0cm]{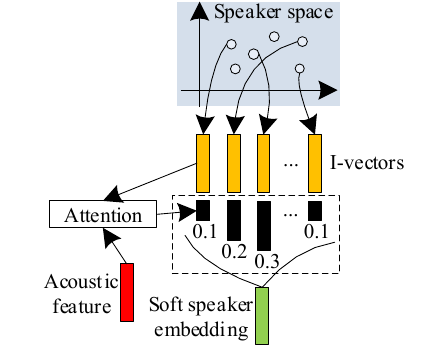}}
 \caption{A schematic representation of the soft speaker embedding}
\label{fig:SAM}
\end{minipage}
\hfill
\begin{minipage}[b]{0.48\linewidth}
  \centering
  \centerline{\includegraphics[width=6.0cm]{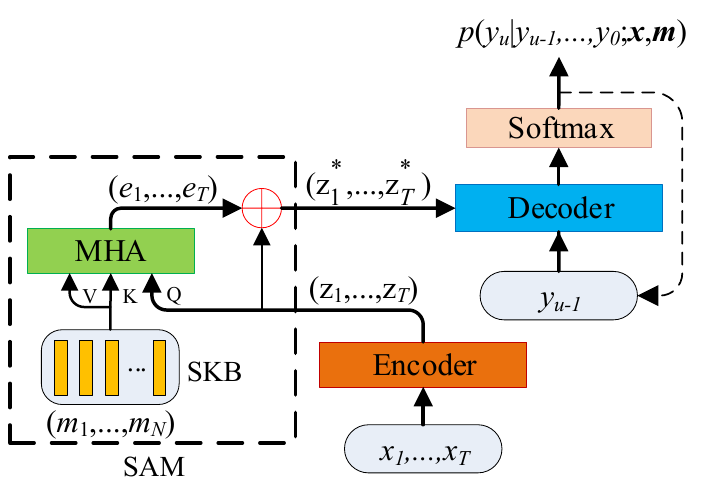}}
\caption{A schematic representation of the proposed speaker aware speech transformer (SAST)}
\label{fig:SAST}
\end{minipage}
%

%
\end{figure*}


\subsection{Speech-Transformer}
\label{ssec:subhead}
Just like most E2E systems \cite{graves2006connectionist, chan2016listen}, Speech-Transformer (ST) has an attention-based encoder-decoder structure. The encoder transforms a speech feature sequence \textbf{x} = $(x_1,...x_T)$ to a continuous representations \textbf{z} = $(z_1,...z_T)$. Given \textbf{z}, the decoder then generates an output sequence \textbf{y} = $(y_1,...y_U)$ one character at a time step.

The encoder of ST is composed of a stack of $N_e$ identical blocks, and each one has two sub-layers. The first sub-layer is a MHA, and the second is a position-wise feed-forward network. Residual connections are employed around each of the two sub-layers, followed by a layer normalization. The decoder composed of a stack of $N_d$ identical blocks is similar to the encoder except inserting a third sub-layer to perform multi-head attention over the output of the encoder stack. To prevent leftward information from flowing and preserve the auto-regressive property in the decoder, all values corresponding to illegal connections are masked in the self-attention sub-layers of decoder. Since no recurrence and no convolution are contained in the model, additional positional encodings are added to the input embeddings at the bottoms of the encoder and decoder stacks to inject some information about the relative or absolute position in the sequence. For more details about ST, we refer the readers to \cite{vaswani2017attention, Zhou2018}.

\section{Proposed Method}
\label{sec:pro}
\subsection{Speaker Embedding: Soft vs. Hard}
\label{subsec:softembed}
The speaker-aware training methods proposed in \cite{saon2013speaker, miao2014towards, abdel2013fast, DBLP:journals/corr/abs-1710-06937} trained the acoustic models with speaker vectors as auxiliary features. These methods are effective to facilitate the models to normalize the speaker variations. However, each speaker vector used in the testing condition is still unseen for the acoustic model. In this work, we treat these speaker vectors as \textbf{hard} speaker embedding, where \textbf{hard} means that for each particular speaker, the speaker representation only depends on the corresponding acoustic information and has no relations with other speakers.


On the contrary, the Speaker Attention Module (SAM) proposed in this paper can produce relatively \textbf{soft} speaker embedding for each speaker. We assume that there are similarities among different speakers and each speaker vector can be represented as a linear combination of a set of basic speaker representations.
Fig. \ref{fig:SAM} gives an illustration of the soft speaker embedding, where i-vectors from the group of basic speakers form a speaker space. Given an acoustic feature vector of one speaker, the similarity of this acoustic vector and each basic i-vector in the speaker space is calculated with the attention mechanism to get the weight for each basic i-vector (the float number under the black bar in Fig. \ref{fig:SAM}). Finally, the soft speaker embedding (the green bar in the figure) is obtained as the weighted sum of the basic i-vectors.

There are mainly two advantages of the soft speaker embedding compared with the hard ones. The first one is generalization. In the decoding condition, the model trained with soft embedding has the ability to get reasonable weighted combination of the basic speaker representations for each unseen speaker. However, for the hard ones, the embedding of unseen speaker is totally new for the trained model. The second advantage is that there is no need to compute the i-vector of each testing speaker for soft embedding model, since only the general acoustic features are needed for decoding. This makes the feature front-end simpler than the hard one.

\subsection{Speaker-Aware Speech-Transformer}
\label{subsec:sast}

The model proposed in this work is Speaker-Aware Speech-Transformer (SAST), which is a standard Speech-Transformer (ST) equipped with a Speaker Attention Module (SAM). The structure of ST is similar to \cite{Zhou2018, yuanyuan2019speech} and we mainly focus on the details of SAM in this part.

Fig. \ref{fig:SAST} shows the architecture of SAST, where the dashed box illustrates the structure of SAM. SAM mainly consists of two parts: the Speaker Knowledge Block (SKB), and the multi-head attention layer. The SKB is a static memory which is made of i-vectors from a group of basic speakers. These i-vectors form the speaker space and are denoted as $\textbf{m}=(m_1,...,m_N)$, where $N$ is the number of i-vectors in the knowledge block. The second part of SAM is a MHA layer, of which the query is the encoder output $\textbf{z}=(z_1,...,z_T)$, and the key and value are the basic i-vectors $\textbf{m}=(m_1,...,m_N)$. The computation of multi-head attention is same as Eq. (\ref{eq2}). To make it clear in SAM, we unfold this process as follows:

For the $i$-th head, the first step is projection:
\begin{equation}
 \setlength{\abovedisplayskip}{4pt}
 \setlength{\belowdisplayskip}{4pt}
  z_t^i = {W}_{q}^iz_t
  \label{eq5}
\end{equation}
\begin{equation}
 \setlength{\abovedisplayskip}{4pt}
 \setlength{\belowdisplayskip}{4pt}
  m_n^i = {W}_{kv}^im_n
  \label{eq6}
\end{equation}
where $z_t$ and $m_n$ are projected to dimension $d_{q}$ and $d_{kv}$ respectively, where $d_{q}$ equals with $d_{kv}$. The second step is to compute the similarity between the encoder output and the speaker vector, with scaled dot-product attention:
\begin{equation}
 \setlength{\abovedisplayskip}{4pt}
 \setlength{\belowdisplayskip}{4pt}
  u_{nt}^i = \text{softmax}(\frac{m_n^iz_t^i}{\sqrt{d_q}})
  \label{eq7}
\end{equation}
where the scalar $1/\sqrt{d_q}$ is used to prevent softmax from falling into regions with very small gradients. Then $u_{nt}^i$ is used to summarize $\textbf{m}$ at time step t:
\begin{equation}
 \setlength{\abovedisplayskip}{4pt}
 \setlength{\belowdisplayskip}{4pt}
  e_t^i = \sum_{n=1}^N{u_{nt}^im_n^i}
  \label{eq8}
\end{equation}
where $e_t^i$ is the soft speaker embedding at time step $t$ for the $i$-th head. Finally, the speaker embedding at time step $t$ is obtained by concatenating $e_t^i$ from $h$ heads:
\begin{equation}
 \setlength{\abovedisplayskip}{4pt}
 \setlength{\belowdisplayskip}{4pt}
  e_t = [e_t^1;...;e_t^h]
  \label{eq9}
\end{equation}

It should be noted that the speaker embedding vector $e_t$ in Eq. (\ref{eq9}) is time step dependent, i.e., there is a different embedding vector for each frame. Another choice is doing at utterance level, that is, generating only one speaker vector for each utterance. We believe that the frame-level embedding is more effective since it's more powerful to capture the variation in one utterance over time, including the speed, mood and tone, etc.

The symbol \textcircled{+} in Fig. \ref{fig:SAST} denotes a concatenate operation. The encoder output and the speaker embedding vector at each time step are attached together:
\begin{equation}
 \setlength{\abovedisplayskip}{4pt}
 \setlength{\belowdisplayskip}{4pt}
  z_t^*=[z_t;e_t]
  \label{eq10}
\end{equation}

Compared to the speaker-independent (SI) ST, the decoder in SAST receives additional soft speaker embedding vectors as inputs, making the model more robust to the speaker variations.
\section{Experiments}
\label{sec:majhead}
We conduct five experiments to investigate the proposed speaker attention module. The first three focus on three factors of SAM: the number of basic i-vectors in the knowledge block, the location of SAM in the encoder, and the level of soft speaker embedding (frame-level or utterance-level). The fourth experiment is designed to compare the performance of soft and hard speaker embedding. In the last one, we study whether or not the i-vectors in the knowledge block have to come from the same dataset used for acoustic training.

\subsection{Data Set}
\label{ssec:subhead}
Experiments are mainly executed with AISHELL-1 data set \cite{DBLP:journals/corr/abs-1709-05522}, which is a Chinese Mandarin speech corpus with $178$ hours data. It is recorded by $400$ speakers from different accent areas in China and is divided into training, development and test set with $340$, $40$ and $20$ speakers respectively, without speaker overlapping. The development and test set are used to evaluate the model performance, which contains $14,326$ utterances (about $10$ hours) and $7,176$ utterances (about $5$ hours), respectively.

The $5$-th experiment and the training of i-vector estimator utilize AISHELL-2\cite{DBLP:journals/corr/abs-1808-10583}. It contains about $1000$ hours Chinese Mandarin speech data, with $1991$ speakers in total. It should be noted that AISHELL-1 is a subset of AISHELL-2 data set.

\subsection{I-vector Estimator}
\label{ssec:subhead}
In order to get high quality i-vectors, the i-vector estimator is trained with all the training data of AISHELL-2 (instead of AISHELL-1). The training process follows the SRE08 recipe in Kaldi toolkit. A 2048 diagonal component universal background model (UBM) is first trained, and then $200$-dimensional i-vectors are extracted and further compressed to $100$ dimension by linear discriminant analysis (LDA) followed by length normalization. 

\subsection{Experimental Setup}
\label{ssec:subhead}
All the acoustic models are trained with $80$-dimensional log-Mel filter-bank features, computed with a $25$\,ms window and shifted every $10$\,ms. The raw features are normalized via mean subtraction and variance normalization per speaker side. Before sending into the transformer, the features are firstly stacked with $3$ frames to the left and then down-sampled to $33.3$ Hz frame rate.

Both the baseline ST and the proposed SAST contain $6$ encoder and $6$ decoder blocks, with a per-block configuration of $d_{model}=512$, $16$ attention heads, and $2048$ feed forward inner-layer dimension. The MHA layer in the speaker attention module of SAST also has $16$ heads, with $d_q$ and $d_{kv}$ being $32$ for each head. There are $4234$ output units in total, including $4230$ Chinese characters, plus $4$ extra tokens, i.e., an unknown token ($<$UNK$>$), a padding token ($<$PAD$>$), and sentence start and end tokens ($<$S$>$/$<$$\backslash$S$>$). During training, the samples are shuffled randomly and then batched together with $256$ batch size. We use the Adam optimizer with ${\beta _1} = 0.9$, ${\beta _2} = 0.98$, $\varepsilon  = {10^{ - 9}}$ and alter the learning rate over the course of training. During training, the label smoothing of value ${\varepsilon _{ls}} = 0.1$ is employed\cite{szegedy2016rethinking}.
All the models are trained for $60$ epochs, and are evaluated every $2$ epochs on the development set of AISHELL-1. The model that performs best on the development set is chosen, and $5$ model checkpoints before it are averaged to get the final model, which is then used to decode the test sets. For evaluation, we use beam search with a beam size of $5$ and length penalty $\alpha = 0.6$. No external language model is used in this work.

The first baseline we consider is chain model, which obtains 7.46\% CER on the test set of AISHELL-1, according to the results in Kaldi toolkit \cite{povey2011kaldi}. The second one is LAS model \cite{Changhao2019improving} achieving $10.56$\% CER (without any external language model) on this task. Our ST baseline get 8.36\% CER, which performs much better than LAS model, meaning that our baseline is competitive.

\begin{figure}[h]
  \vspace{2mm}
  \centering
  \includegraphics[width=6cm,height=3.5cm]{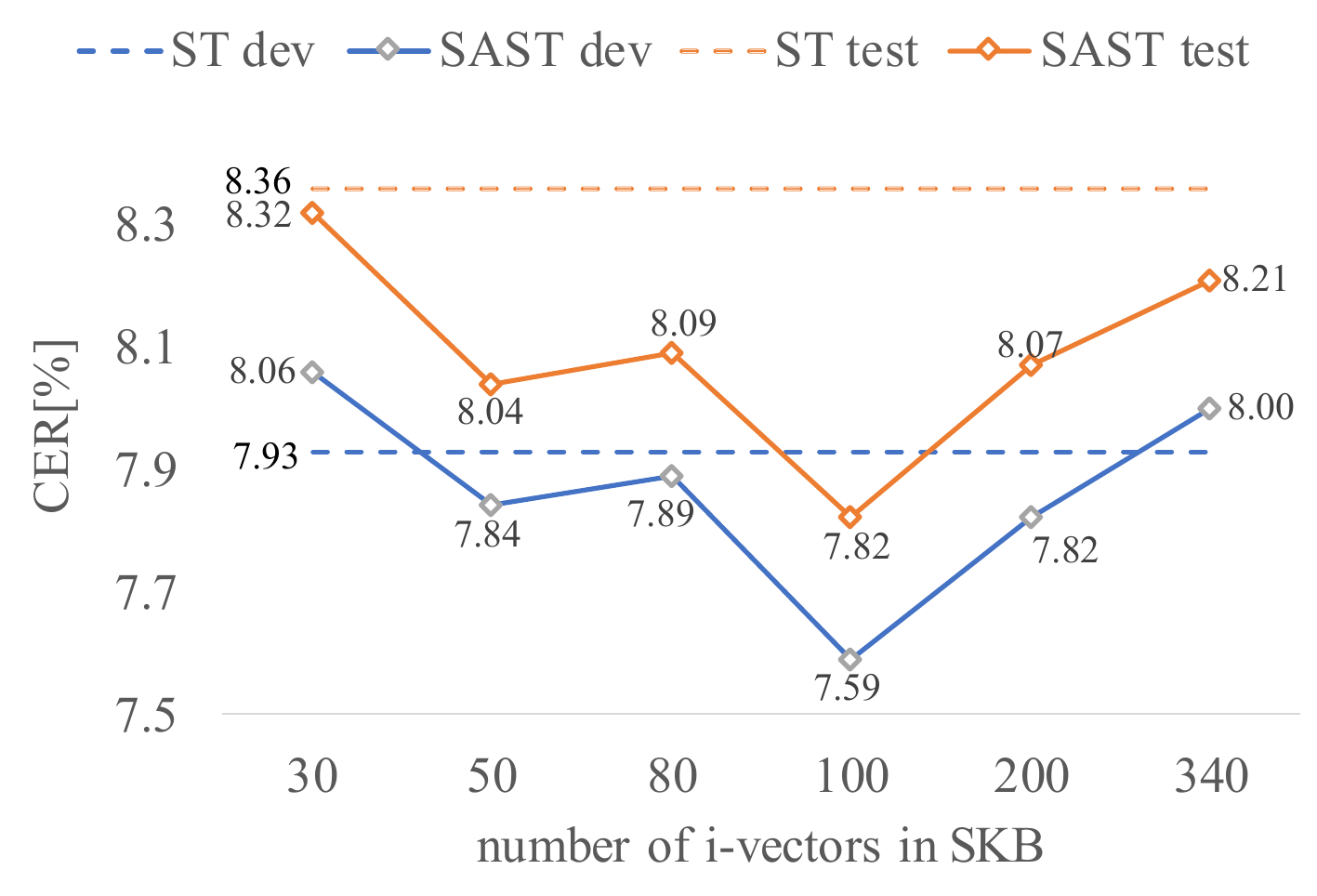}
  \caption{CER[\%] of ST baseline and SAST models with different numbers of i-vectors in SKB on development and test set of AISHELL-1.}
  \label{fig:2}
\end{figure}
\subsection{Experimental Results}
\label{ssec:ER}

\subsubsection{The Number of Basic I-vectors in SKB}
\label{sssec:ivector-num}
The first problem about SAM is how many basic i-vectors there should be in the speaker knowledge block (SKB). We randomly select different numbers of speakers from the training set of AISHELL-1 (totally $340$ speakers), ranging from $30$ to $340$, extract the corresponding i-vectors and put them in the SKB. Several SAST models with different SKB are trained with the same whole AISHELL-1 training set. Please note that gender balancing is considered when choosing these speakers, i.e., half of speakers are male and half are female.

For experiments in this subsection, the speaker attention module is placed on the highest block (the $6$th block) of the encoder in SAST, that is, $z_t$ in Eq. (\ref{eq5}) is the output of the $6$-th encoder block. Besides, the soft speaker embedding is performed at the frame-level, just following Eq. (\ref{eq8}).

Performance of different SAST models, as well as the baseline ST, is presented in Fig. \ref{fig:2}. It shows clearly two U-shaped CER curves for the development and test set when the number of i-vectors in SKB is increased from $30$ to $340$. The best SAST model is obtained when $100$ i-vectors are provided in SKB, with CER of $7.59$ on development set and CER of $7.82$ on test set, which is $4.3\%$ and $6.5\%$ relative CER reduction over baseline ST.

When SKB contains too few i-vectors, e.g. only $30$, the SAST model performs worse than the baseline ST. It's easy to understand since these few i-vectors can not represent the speaker space quite well. On the other hand, the performance of SAST is also not good when all the $340$ i-vectors of AISHELL-1 training data are offered to SKB. This is because for each training utterance, there exists a corresponding speaker i-vector in SKB and the SAM can easily find it. The model has no opportunity to learn how to obtain the soft speaker embedding for unseen speakers by combining the basic i-vectors, which can limit the model's generalization. By contrast, when SKB has only $100$ i-vectors, the model is forced to generate suitable soft speaker embedding for acoustic frames belonging to the rest $240$ speakers during training, making the model generalize better for unseen speakers.
\begin{table}[t]
  \caption{CER[\%] of SAST models with different locations of SAM. Gains are relative to our ST baselilne on test set.}
 \vspace{2mm}
  \label{tab:2}
  \centering
  \begin{tabular}{cccc}
    \toprule
    Layer & Dev & Test&Gain \\
      \hline
         2-th & 8.02 & 8.30 & 0.7\%\\
         4-th & 7.97 & 8.13 & 2.8\% \\
         6-th & \textbf{7.59} & \textbf{7.82} & \textbf{6.5\%} \\
    \bottomrule
  \end{tabular}
\end{table}
\subsubsection{The Location of SAM in the Encoder}
\label{sssec:location}

The second factor that we focus on is the location of SAM in the encoder. In our previous experiments of Sec.\ref{sssec:ivector-num}, the speaker attention module is located at the $6$-th (highest) block of encoder in SAST model. In this subsection, another two SAST models are trained with SAM settled on the $2$-th or the $4$-th encoder block, respectively. The implementation details of SAM are slightly different when located at lower blocks. The query vector $z_t$ in Eq. (\ref{eq5}) is the output of lower block, but when concatenating with the speaker vector $e_t$ in Eq. (\ref{eq10}), $z_t$ is the output of highest encoder block instead of the lower one. By doing so, the speaker embedding is expected to influence the decoder more directly.

The number of i-vectors in SKB is $100$, and the speaker embedding is performed at the frame-level, following the Sec.\ref{sssec:ivector-num}. Results are shown in Table~\ref{tab:2}. It's clear that higher level acoustic representation is more helpful for SAM to calculate the soft speaker embedding vector. It's in line with our expectations since higher encoder block outputs more abstract features. In the following experiments, the speaker attention module is located at the highest block of encoder.

\subsubsection{The Level of Speaker Embedding in SAM}
\label{sssec:level}

In this part, we investigate whether we should obtain the speaker embedding vector per frame or per utterance. We argue that frame-level embedding is more effective since it can capture the variance over time in one utterance in Sec.\ref{subsec:sast}. All our previous experiments are executed with frame-level speaker embedding. Now we will compare the performance of these two embedding methods.

Eq. (\ref{eq5}) $\sim$ Eq. (\ref{eq10}) in the Sec.\ref{subsec:sast} show the process of speaker embedding at the frame level. As for utterance-level embedding, we average the output of encoder along the time steps, thus $z_t$ in Eq. (\ref{eq5}) now becomes:
\begin{equation}
 \setlength{\abovedisplayskip}{3pt}
 \setlength{\belowdisplayskip}{3pt}
  z=\frac{1}{T}\sum_{t=1}^{T} {z_t}
  \label{eq11}
\end{equation}
where $T$ is the total time steps in one utterance. $z$ is time step independent and is used to compute the utterance-level speaker embedding for one utterance.

We tried three different settings for the number of i-vectors in SKB: $50$, $100$, and $200$. The results are shown in Table \ref{tab:3} (CER of frame-level models comes from Fig. \ref{fig:2}). It's obvious that the utterance-level speaker embedding performs worse than the frame-level one for all the three settings. This is consistent with our intuition. Speaker's mood, speed or tone can not keep constant over time in one utterance. Frame-level speaker embedding is able to calculate a reasonable speaker vector according to the acoustic features of current time step, thus it's more effective than utterance-level embedding to catch the variations over time.
\begin{table}[t]
  \caption{CER[\%] for frame-level and utterance-level soft speaker embedding on test set of AISHELL-1.}
 \vspace{2mm}
  \label{tab:3}
  \centering
  \begin{tabular}{cccc}
    \toprule
     \#I-vectors & frame-level  & utterance-level \\
      \hline
         50 & 8.04 & 8.23  \\
         100 & 7.82 & 8.14 \\
         200 & 8.07 & 8.23  \\
    \bottomrule
  \end{tabular}
\end{table}
\subsubsection{Hard and Soft Speaker Embedding}
\label{sssec:hvss}

We discussed soft and hard speaker embedding in Sec.\ref{subsec:softembed} and hold the view that soft speaker embedding performed by SAM is more effective than the hard one. In this subsection, we compare them quantitatively.

For soft speaker embedding (SAST model), we choose the best settings according to the findings in previous three experiments: the number of i-vectors in SKB is $100$, SAM is located at the highest encoder block, and speaker embedding is performed at the frame-level. As for the hard embedding, two approaches are tried. The first method is named as \emph{input of ST \textcircled{+} i-vector} (in Table \ref{tab:4}), which concatenates the i-vector for a given speaker to every frame belonging to that speaker, and sends the resulting feature to the input of ST model. The second one is called \emph{output of encoder \textcircled{+} i-vector}, in which the output of highest encoder block at each time step is spliced with the i-vector of the corresponding speaker.
\begin{table}[h]
  \caption{CER[\%] for ST, SAST with $100$ i-vectors in AISHELL-1 and attaching i-vector to model (input or output of encoder) directly. Gains are relative to our ST baselilne on test set.}
 \vspace{2mm}
  \label{tab:4}
  \centering
  \setlength{\tabcolsep}{1mm}
  \begin{tabular}{cccc}
    \toprule
    Models & Dev &Test  & Gain\\
     \midrule
         ST baseline & 7.93 & 8.36 & - \\
         + input of ST \textcircled{+} i-vector& 7.85 & 8.22 & 1.7\% \\
         + output of encoder \textcircled{+} i-vector& 8.21 & 8.52 & - \\
         SAST & \textbf{7.59} & \textbf{7.82} &\textbf{6.5\%}\\
    \bottomrule
  \end{tabular}
\end{table}

According to Table \ref{tab:4}, both of the two hard speaker embedding methods perform worse than the soft one, i.e., SAST model proposed in this work. Compared with hard embedding, the soft speaker embedding performed by SAM empowers SAST to generalize better to unseen speakers.
\begin{table}[th]
  \caption{CER[\%] for SAST with access to different number of i-vectors in AISHELL-2. Gains are relative to our ST baselilne on test set.}
 \vspace{2mm}
  \label{tab:5}
  \centering
  \begin{tabular}{cccc}
    \toprule
     {\#I-vectors} &Dev & Test &Gain \\
      \hline
          50 &7.75 &8.21 & 1.8\% \\
         100 & 7.80 & 8.19 & 2.0\% \\
         200 & 7.87 & 8.19 & 2.0\%\\
         300 & \textbf{7.78} & \textbf{7.97} & \textbf{4.7\%}\\
         400 & 7.79 & 8.15 & 2.5\%\\
    \bottomrule
  \end{tabular}
\end{table}
\subsubsection{Source of I-vectors in SKB}
In aforementioned experiments, i-vectors in the SKB are extracted from speakers in the training set of AISHELL-1, the acoustic data of them is also used to train SAST models. Thus, there are overlaps of speakers used in SKB and SAST training. In this subsection, we are interested in a question, that is, whether or not the i-vectors in the knowledge block have to come from the same dataset used for acoustic model training. We design the following experiments with AISHELL-2 as the additional resource. First we exclude the $400$ speakers of AISHELL-1 from all the $1991$ speakers in AISHELL-2 (noting that AISHELL-1 is a subset of AISHELL-2). Then we randomly select different numbers of speakers from the remaining $1591$ speakers of AISHELL-2 (gender balancing is also considered), extract the corresponding i-vectors, and put them in SKB. All the SAST models are trained with the training set of AISHELL-1, so there are no speaker overlaps in SKB and acoustic model training.

Model performances on the development and test set of AISHELL-1 are presented in Table \ref{tab:5}. We can see that the proposed SAST still works quite well even if the i-vectors in SKB all comes from a different data source other than the acoustic training set. We think that this can be attributed to the power of soft speaker embedding. The i-vectors in SKB form a speaker space, and the SAST model is trained to represent each speaker as the combination of data points in this space, even for speakers not belonging to SKB.
\section{Conclusions}
\label{sec:conclusions}
In this paper, we propose a model called Speaker-Aware Speech-Transformer (SAST), which is a standard speech-transformer equipped with speaker attention module (SAM). SAM has a speaker knowledge block (SKB) that is a group of i-vectors, and is trained to produce a soft speaker embedding for each acoustic frame, which helps the SAST model to normalize the speaker variations. We investigate different factors of SAM: the number of i-vectors in SKB, the location of SAM in the model, and the level of soft speaker embedding. On AISHELL-1 test set, the proposed SAST gives a 6.5\% relative CER reduction over the baseline ST. What's more, we also demonstrate that soft speaker embedding is superior to the hard one, and the i-vectors in SKB do not have to come from the same dataset used for acoustic training.



\pagebreak

\bibliographystyle{IEEEbib}
\bibliography{strings,refs}

\end{document}